\documentclass{article}
\usepackage{tccml_iclr2025_conference, times}
\usepackage{hyperref}
\usepackage{tikz}
\usepackage{graphicx, subcaption}
\usepackage{booktabs}
\usepackage{amsfonts}
\usepackage{float}
\usepackage{wrapfig}
\usepackage[flushleft]{threeparttable}
\usepackage{lipsum}
\usetikzlibrary{shapes.geometric, arrows, positioning}
\iclrfinalcopy
\definecolor{lightgreen}{rgb}{0.8, 0.9058, 0.8117}
\definecolor{lightblue}{rgb}{0.7607, 0.9098, 0.9686}
\definecolor{coral}{rgb}{0.9882, 0.8784, 0.8823}
\tikzstyle{box} = [rectangle, minimum width=3cm, minimum height=0.5cm, text centered]
\tikzstyle{rounded} = [rectangle, rounded corners, minimum width=2.5cm, minimum height=0.8cm, text centered, draw=black, line width=0.2mm, fill=coral]
\tikzstyle{rounded_g} = [rectangle, rounded corners, minimum width=2.5cm, minimum height=0.8cm, text centered, draw=black, line width=0.2mm, fill=lightgreen]
\tikzstyle{rounded_b} = [rectangle, rounded corners, minimum width=2.5cm, minimum height=0.8cm, text centered, draw=black, line width=0.2mm, fill=lightblue]
\tikzstyle{FFNN} = [trapezium, trapezium angle=60, draw = black, line width=0.2mm,  minimum width=2.5cm]
\tikzstyle{observation} = [circle]
\tikzstyle{arrow} = [thick,->,>=stealth, line width=0.3mm]
\tikzstyle{concatnode} =[circle, draw, inner sep=0pt, line width=0.2mm]

\title{Drought forecasting using a hybrid neural\\ architecture for integrating time series \\ and static data}

\author{Julian Agudelo  \thanks{julian.agudeloacosta@agroparistech.fr, j.agudelo@agrial.com}\\
AgroParisTech, Agrial\\
\And
Vincent Guigue \thanks{vincent.guigue@agroparistech.fr}\\
AgroParisTech\\
\And
Cristina Manfredotti \\
AgroParisTech\\
\And
Hadrien Piot \\
Agrial\\
}
\date{January 2025}

\begin{document}

\maketitle

\begin{abstract}
Reliable forecasting is critical for early warning systems and adaptive drought management. Most previous deep learning approaches focus solely on homogeneous regions and rely on single-structured data. This paper presents a hybrid neural architecture that integrates time series and static data, achieving state-of-the-art performance on the DroughtED dataset. Our results illustrate the potential of designing neural models for the treatment of heterogeneous data in climate related tasks and present reliable prediction of USDM categories, an expert-informed drought metric. Furthermore, this work validates the potential of DroughtED for enabling location-agnostic training of deep learning models. All the necessary code to reproduce the experiments is available at \url{https://github.com/JulAgu/drought_forecasting_HM}.
\end{abstract}

\section{Introduction}
\label{introduction}
Drought is a natural phenomenon characterized by a prolonged period of below-average precipitation, which causes significant hydrological imbalances that negatively impact land resources \citep{reichhuber2023}. Droughts are typically classified into meteorological, agricultural, hydrological and socioeconomic types \citep{humphreys1931, dracup1980, wilhite2007}, reflecting their broad impacts on different systems. As global warming intensifies, drought events' frequency, duration, and severity increase, exacerbating vulnerabilities. Given the multifaceted nature of droughts and their effects across various spatial and temporal scales, reliable forecasting is crucial for early warning systems and adaptive resources management. Effective predictions help mitigate impacts on water supply, agriculture, ecosystems, and communities \citep{aghelpour2020}.

Several indices are commonly used to assess drought conditions. Some well-known examples are the Palmer Drought Severity Index (PDSI) \citep{palmer1968}, the Standardized Precipitation Index (SPI) \citep{mckee1993}, and the Standardized Precipitation Evapotranspiration Index (SPEI) \citep{vicente-serrano2010}. These indices depend on meteorological variables to quantify deviations from climatic norms. Alternatively, the U.S. Drought Monitor (USDM) categories provide a complete evaluation by integrating hydrological, climatic and weather data with experts' insights, capturing a holistic view of drought impacts \citep{svoboda2002}. We use the USDM categories as our target variable for this study due to their comprehensive nature and integration into DroughtED \citep{minixhofer2021}.

Various authors have used machine learning to predict drought indices \citep{nandgude2023}. Traditionally, manual feature extraction has been used to feed classical machine learning algorithms \citep{gaikwad2015}. However, in recent years, there has been a significant shift toward deep learning approaches, which employ representation learning to automatically extract meaningful features from data \citep{zheng2018}. A considerable number of prior studies utilizing deep learning techniques for drought prediction have predominantly focused on homogeneous regions \citep{dikshit2022}, and primarily employed uniform structured data, as images \citep{chaudhari2023} or time series \citep{lalika2024, vijaya2023} alone.

This paper uses the DroughtED dataset introduced by \cite{minixhofer2021} and presents a novel modeling approach for drought forecasting with heterogeneous data. The contributions of this research are threefold. (1) We introduce a neural architecture integrating time series and static data through FFNNs, LSTMs, categorical embeddings, and an attention mechanism. We benchmark our model against the dataset's baselines. (2) We conduct an ablation study to assess the contribution of each component within the proposed model. (3) We apply visualization techniques over latent states to perform model introspection.

\section{Data}
\label{data}
DroughtED is a large-scale dataset designed to forecast drought conditions in the United States by integrating spatial and temporal features \citep{minixhofer2021}.It includes historical meteorological time series, soil physical characteristics, and historical drought intensity information at the county level. The meteorological data are sourced from The NASA Prediction Of Worlwide Energy Resources (NASA POWER) project \citep{zhang2009}, the soil properties are derived from the Harmonized World Soil Database \citep{nachtergaele2008} and the drought intensity evaluations are taken from the USDM \citep{svoboda2002}.

Drought data are ordinal indicators measured locally. These indicators are then reduced to continuous average values and aggregated at the county level. The target values to be predicted are 6 continuous values at county level, corresponding to 6 consecutive weeks. We have data $\mathbf{Y}\in \mathbb R^{N\times 6}$ where $N$ corresponds to the number of pairs (county, timestamp) noted $(c,t)$ in the following. To predict these targets, we have static descriptors $\mathbf{S} \in \mathbb{R}^{C \times f}$, where $f$ is the number of features describing the soil physical properties, and $C$ the number of counties. Note that the descriptors will be divided into categorical $\mathbf{s_d} \in \mathbb R^{f_d}$ and numerical $\mathbf{s_n} \in \mathbb R^{f_n}$ features below. The meteorological data is represented as multivariate time series grouped in a tensor $\mathbf{X} \in \mathbb{R}^{C \times P \times M}$: for each county, we observe $M=20$ different measurements over several years corresponding to $P$ days.

We use all the descriptors from \citep{minixhofer2021}: each local target to predict $\mathbf{y} \in \mathbb R^6$ (6 weeks following $t$ in county $c$) is associated with static descriptors $\mathbf{s}_d$ and $\mathbf{s}_n$ as well as a multivariate time series $\mathbf{x} \in \mathbb R^{T\times M'}$. The period extracted from $\mathbf{X}$ corresponds to $T=180$ days before the timestamp $t$ for the county $c$, while we take the $M$ available measurements plus the $M$ measurements corresponding to the previous year over the same days to enable the model to build comparative features, thus $M'=2M$.

Counties are indexed using the FIPS \citep{FIPS1990} identifier. We use the train, validation and test splits available in Kaggle\footnote{\url{https://www.kaggle.com/datasets/cdminix/us-drought-meteorological-data}}.

\section{The proposed model}
\label{model}

\begin{wrapfigure}[19]{L}{0.49\textwidth}
\vspace{-0.4cm}
    \centering
    \begin{tikzpicture}[node distance=1.5cm, scale=0.69, transform shape]
        \node (Output) [box] {Output};
        \node (MLP_Combined) [FFNN, fill=coral, below of=Output] {MLP};
        \node (concat_tab) [concatnode, below of=MLP_Combined, yshift=0.1cm] {\huge $\mathbf{\oplus}$};

        \node (C) [box, below of = concat_tab, yshift=-4.5cm] {Time series};
        \node (B) [box, right of = C, xshift=1.8cm] {Numerical features};
        \node (A) [box, left of = C, xshift=-2cm] {Categorical features};
        
        \node (F) [rounded_g, above of=C] {LSTM};
        \node (G) [rounded_g, above of=F, yshift=0.2cm] {Attention};
        \node (H) [rounded_g, above of=G, yshift=-0.3cm] {Context vector};

        \node (D) [rounded_b, above of=A] {Embeddings};
        \node (FFNN_emb) [FFNN, fill=lightblue, above of=D, yshift=0.2cm] {FFNN};
   
        \draw (A.north) node [left, yshift=0.3cm] {$\mathbf{s}_d$};
        \draw (D.north) node [left, yshift=0.3cm] {$\mathbf{e}$};
        \draw(FFNN_emb.north) node [left, yshift=0.3cm] {$\mathbf{e'}$};
        
        \draw ([xshift=-0.6cm]C.north) node [left, yshift=0.3cm] {$x_{1}$};
        \draw ([xshift=0.6cm]C.north) node [right, yshift=0.3cm] {$x_T$};
        \draw (F.north) node [above] {$\cdots$};
        \draw ([xshift=-0.6cm]F.north) node [left, yshift=0.2cm] {$h_{1}$};
        \draw ([xshift=0.6cm]F.north) node [right, yshift=0.2cm] {$h_{T}$};
        \draw (C.north) node [above] {$\cdots$};
        \draw(H.north) node [left, yshift=0.3cm] {$\mathbf{\tilde{h}}$};
        \draw (B.north) node [left, yshift=0.3cm] {$\mathbf{s}_n$};
        \draw (concat_tab) node [left, yshift=0.55cm] {$\mathbf{x'}$};
        \draw (MLP_Combined.north) node [left, yshift=0.3cm] {$\hat{y}$};

        \draw [arrow] (A) -- (D);
        \draw [arrow] (H) -- (concat_tab);
        \draw [arrow] (D) -- (FFNN_emb);
        \draw [arrow] (FFNN_emb) |- (concat_tab);

        \draw [arrow] ([xshift=-0.6cm]C.north) -- ([xshift=-0.6cm]F.south);
        \draw [arrow] ([xshift=0.6cm]C.north) -- ([xshift=0.6cm]F.south);
        \draw [arrow] ([xshift=0.6cm, yshift=-0.4cm]G.south) -- ([xshift=1.5cm, yshift=-0.4cm]G.south) |- (concat_tab.south east);
        \draw [arrow] (B) |- (concat_tab.north east);

        \draw [arrow] ([xshift=-0.6cm]F.north) -- ([xshift=-0.6cm]G.south);
        \draw [arrow] ([xshift=0.6cm]F.north) -- ([xshift=0.6cm]G.south);
        \draw [arrow] (G) -- (H);
        
        \draw [arrow] (concat_tab) -- (MLP_Combined);
        \draw [arrow] (MLP_Combined) -- (Output);
    \end{tikzpicture}
    \caption{Schematic view of the proposed model.}
    \label{fig_app:model_arch}
\end{wrapfigure}
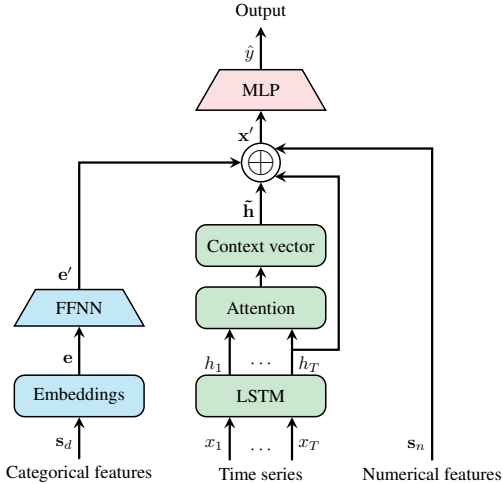

Despite the flexibility of neural network libraries, handling data with heterogeneous structures is still an open topic in deep learning \citep{guo2019, kamm2023}. We propose a hybrid neural model that combines four modules: Long-Short-Term Memory Recurrent Neural Networks (LSTM), Feedforward Neural Networks (FFNN), embedding layers and an attention mechanism. Figure \ref{fig_app:model_arch} shows the schema of the proposed model.

The categorical features $\mathbf{s_d}$ are passed through the embedding layer $E$, which maps them to dense vectors $\mathbf{e} \in \mathbb{R}^{z}$.
Then, the dense representations of all categorical features are concatenated and passed through a FFNN $\mathcal{F}$ that reduces its dimensionality, resulting in a vector $\mathbf{e'} = \mathcal{F}(E(\mathbf{s_d})) \in \mathbb{R}^{z'}$, where $z'<z$.
In parallel, the multivariate time series $\mathbf{x}$ is fed into the LSTM, yielding hidden states at each time step $t$: $\{\mathbf{h}_t\}_{t=1:T} \in \mathbb{R}^{h\times T}$. The hidden states are further processed by the attention mechanism (detailed in Appendix \ref{app:attention}) to produce a context vector $\mathbf{\tilde h} \in \mathbb{R}^h$. The numerical features $\mathbf{s_n} \in \mathbb{R}^{f_n}$ remain unchanged.

The final representation is obtained by concatenating the context vector $\mathbf{\tilde h}$, the last hidden state $\mathbf{h}_T$, the continuous features $\mathbf{s_n}$, and the latent representation of the categorical features $\mathbf{e'}$.

The resulting vector is $\mathbf{x'} = [\mathbf{\tilde h}, \mathbf{h}_T, \mathbf{e'}, \mathbf{s}_n] \in \mathbb{R}^{2h + z' + f_n}$. This concatenated vector is then passed through an MLP $\mathcal{M}$, which outputs the prediction $\hat{y} = \mathcal{M}(\mathbf{x'}) \in \mathbb R^6$.

To the best of our knowledge, no prior effort has employed a neural architecture combining these elements to predict USDM drought categories.

\setlength{\columnsep}{0.3cm}

\section{Experiments and results}
Given the outlined framework, we conducted some experiments to investigate the following questions. a) How does the proposed hybrid architecture perform compared to the best baseline model established by \cite{minixhofer2021} regarding predictive performance and generalization? b) What is the relative contribution of each architectural component to the overall performance? c) Does the proposed model sustain superior performance under a location-agnostic training compared to location-specific training? d) How do the attentional mechanism and latent embeddings interact to shape the learned representations within the model?

Hyperparameter optimization was conducted using Bayesian optimization via Optuna \citep{optuna2019}, with the AdamW optimizer \citep{loshchilov2019} and a cyclical learning rate schedule \citep{smith2017}. We replicated the baseline LSTM model following the publicly available code and guidelines provided by \cite{minixhofer2021}, and we present results from our implementation to ensure consistency. The Transformer baseline model was excluded from the analysis since the LSTM demonstrated superior performance. This decision aligns with the findings of \cite{minixhofer2021} and is further supported by \cite{zeng2023}, which highlights the limitations of transformer-based approaches in time series forecasting.

\paragraph{a) Predictive performance and generalization.}
After training the architecture with the optimal set of hyperparameters, we observed that the proposed Hybrid Model (HM) consistently yields better MAE and macro $F_1$ scores over the weeks, as shown in Table \ref{tab:results_test}. Over the entire test set, the model shows relative improvements of 30\% in the MAE, 9\% in the $F_1$ and 7\% in the multi-class weighted ROC-AUC score compared to the baseline LSTM. To estimate the expected prediction error more accurately, we performed 5-fold cross-validation (Appendix \ref{app:tab:cv_results}) and conducted a paired t-test to evaluate whether HM significantly outperforms the LSTM. The results show a significant improvement in MAE, RMSE, and $F_1$, with \textit{p}-values of 0.03, 0.04, and 0.02, respectively.

\begin{table}[t]
\caption{Weekly results on the test set for the proposed model and the \cite{minixhofer2021} LSTM model.}
\label{tab:results_test}
\begin{center}
    \footnotesize
    \begin{tabular}{l l l l l l l l l l l l l}
    \toprule
    {} & \multicolumn{2}{l}{\textbf{Week 1}} & \multicolumn{2}{l}{\textbf{Week 2}} & \multicolumn{2}{l}{\textbf{Week 3}} & \multicolumn{2}{l}{\textbf{Week 4}} & \multicolumn{2}{l}{\textbf{Week 5}} & \multicolumn{2}{l}{\textbf{Week 6}} \\
    \cmidrule{2-13}
    Model & MAE & F\textsubscript{1} & MAE & F\textsubscript{1} & MAE & F\textsubscript{1} & MAE & F\textsubscript{1} & MAE & F\textsubscript{1}& MAE & F\textsubscript{1} \\
    \midrule
    LSTM & 0.150 & 81.6 & 0.229 & 71.6 & 0.286 & 64.5& 0.347 & 57.4 & 0.394 & 54.2 & 0.432 & 49.6 \\
    HM & \textbf{0.126} & \textbf{82.2} & \textbf{0.169} & \textbf{74.7} & \textbf{0.209} & \textbf{68.6} & \textbf{0.244} & \textbf{64.0} & \textbf{0.269} & \textbf{58.6} &\textbf{0.294} & \textbf{51.0} \\
    \bottomrule
    \end{tabular}
    \vspace{-0.3cm}
\end{center}
\end{table}

\paragraph{b) Ablation study.}
We evaluated the model under three ablation variables (Table \ref{tab:ablation_study}). The best performances are achieved when the attention mechanism was included. Additionally, the ablation study founds that most of the knowledge is derived from meteorological data, which aligns with previous findings \citep{minixhofer2021}.

\begin{table}[t]
\caption{Results of the ablation study for the proposed model (HM).}
\label{tab:ablation_study}
\begin{center}
    \footnotesize
    \begin{tabular}{c c c p{1cm} p{1cm} p{1cm} }
    \toprule
    \multicolumn{3}{c}{Ablation settings} & \multicolumn{3}{c}{}\\
    \cmidrule{1-3}
    Static features & Time series & Attention mech. & MAE & RMSE & F1 \\
    \midrule
    \checkmark & \checkmark & \checkmark & \textbf{0.217} & \textbf{0.377} & \textbf{66.3} \\
    & \checkmark & \checkmark & 0.267 & 0.419 & 56.2 \\
    & \checkmark &  & 0.271 & 0.420 & 56.6 \\
    \checkmark & \checkmark &  & 0.280 & 0.427 & 57.1 \\
    \checkmark &  &  & 0.755 & 0.920 & 21.2 \\
    \bottomrule
    \end{tabular}
\vspace{-0.3cm}
\end{center}
\end{table}

\paragraph{c) Location-agnostic vs location-specific training.}
As in the experiment carried out by \cite{minixhofer2021} for the baseline LSTM, we consider the same 3 states obtained at random (Iowa, Montana and Oklahoma) and trained HM on each state alone and on all training data (Appendix \ref{app:location_expe}). The model trained on data from all counties ---location-agnostic--- demonstrated an average relative improvement of 9.3\%. This indicates that when using HM, location-agnostic training outperforms location-specific training. In comparison, \cite{minixhofer2021} reported an average relative improvement of 4.6\% for the baseline LSTM.

\begin{figure}[h]
    \begin{subfigure}[b]{0.44\textwidth}
    \includegraphics[width=\linewidth]{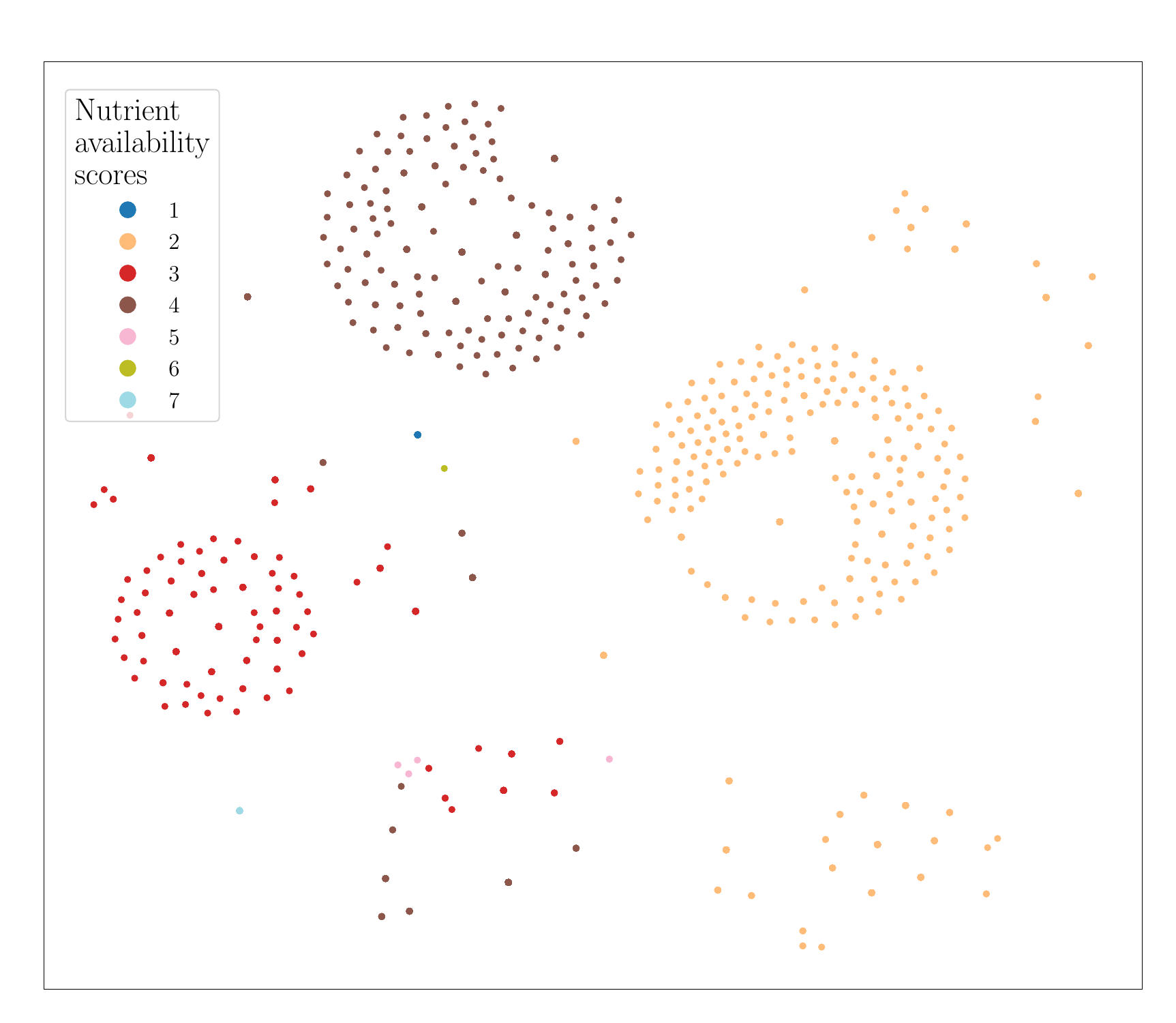}
    \caption{t-SNE colored by ``Nutrient availability''.}
    \label{fig:t-sne}
    \end{subfigure}
\hfill
    \begin{subfigure}[b]{0.46\linewidth}
    \includegraphics[width=\linewidth]{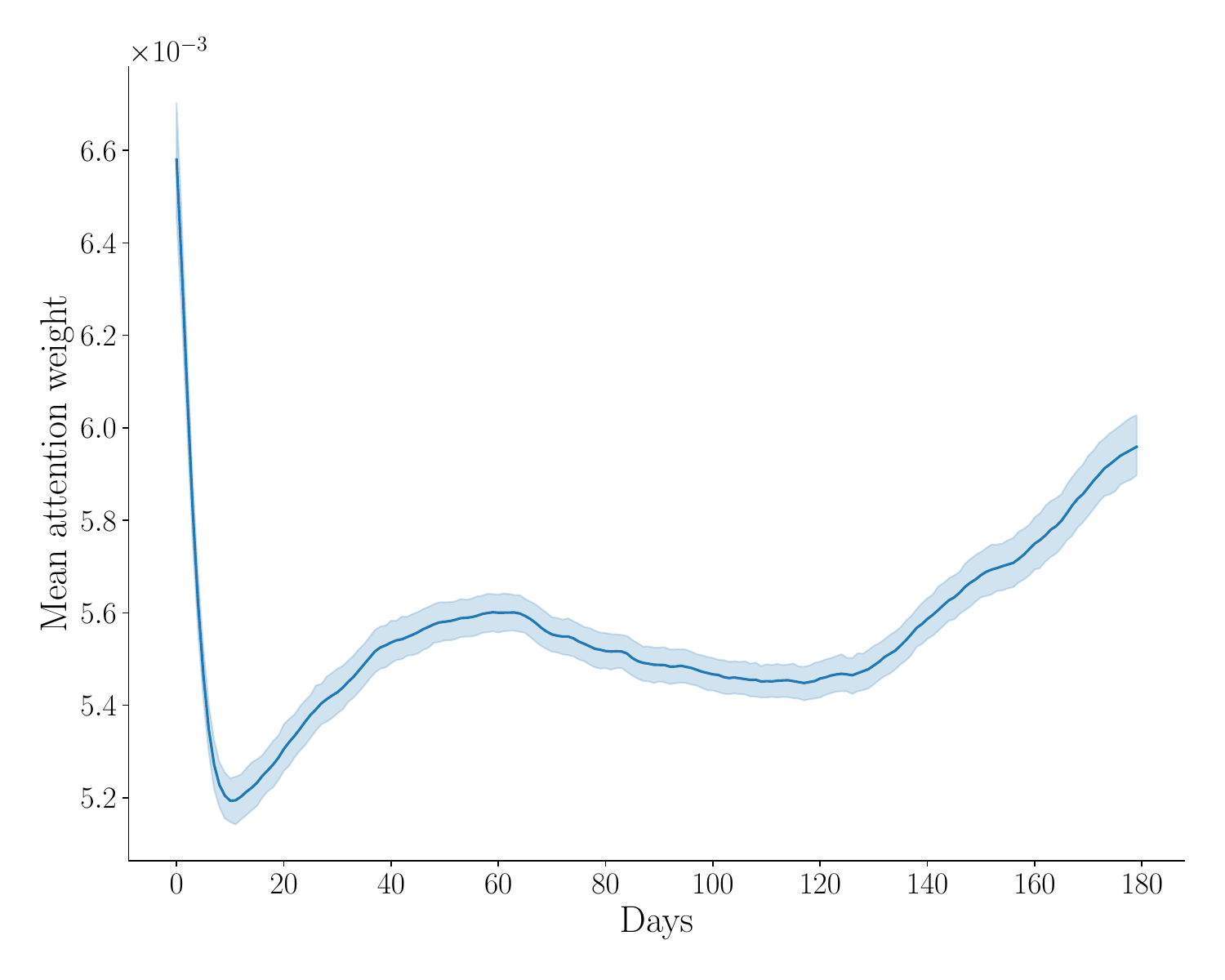}
    \caption{Mean attention weights on the test set.}
    \label{fig:att}
    \end{subfigure}
\caption{t-SNE over the embeddings and mean attention weights curve.}
\label{fig:t-sne_att}
\vspace{-0.3cm}
\end{figure}

\paragraph{d) Model introspection.}
We conduct a qualitative analysis of the model's intermediate representations. By using t-SNE dimensionality reduction (with a perplexity = 100 and 1000 iterations) \citep{vandermaaten2008}, we examined how observations cluster according to each categorical feature. Our findings indicate that the embeddings closely align with the categories of the ``Nutrient availability'' feature (see Figure \ref{fig:t-sne}). We consider this a favorable outcome, as previous work have found that droughts significantly influence the presence and accessibility of nutrients in the soil \citep{he2014, bista2018}.

In relation to the attentional weights, we processed the test set through the model. Then, we plotted the average attention weight for each day, along with a 95\% confidence interval (Figure \ref{fig:att}). Attention is primarily concentrated on the first 10 days, with additional focus on the last 30 days of the look-back window.

\section{Conclusion and future work}

We present a hybrid neural architecture and validate its effectiveness through empirical evaluation on the DroughtED dataset, achieving state-of-the-art performance in forecasting USDM drought categories.

Future improvements to the proposed architecture will prioritize refining the attention mechanism through two strategies: (1) calibrating attention weights sharpness via a learnable softmax temperature parameter and (2) leveraging expert-annotated labels to supervise attention training through an auxiliary loss, mirroring the expert-guided process behind USDM categories. Beyond immediate performance gains, the architecture’s modular design creates a flexible framework for extreme weather events forecasting and other tasks depending on static and temporal data interaction.

\bibliographystyle{iclr2025_conference}
\bibliography{biblio.bib}
\newpage

\appendix

\section{Detail on the attention mechanism}
\label{app:attention}
Since the proposed model is designed as a flexible framework, a simple implementation of the attention mechanism is used. This implementation resembles Bahdanau attention \citep{bahdanau2014}. While their implementation aims to calculate the alignment scores between the sequences of an encoder and a decoder, the proposed approach focuses on differentially weighting the hidden states from the LSTM.

Given the set of hidden states $H = [h_1,h_2,...,h_T]$ at the output of the LSTM, we calculate the scores using a linear layer:

\begin{equation}
    s_t = Wh_t + b
    \label{eq:couche_linaire_attention}
\end{equation}

where $W$ and $b$ are learnable matrix of weights and bias.

Then we calculate the attention weights by passing the scores through a softmax function:

\begin{equation}
    \alpha_t= {softmax}(s_t) = \frac{e^{s_t}}{\sum_{i=1}^{T} e^{s_i}}
    \label{eq:softmax}
\end{equation}

the context vector  $\mathbf{\tilde h}$ is computed as the weighted sum of the LSTM hidden states, using the attention weights.

\begin{equation}
    \mathbf{\tilde h}= \sum_{t=1}^{T} \alpha_{t} h_t
    \label{eq:context_LSTM}
\end{equation}

\section{Selected hyperparameters for each model}
\label{app:hyper}
\begin{table}[h]
\caption{Hyperparameters for the baseline models \citep{minixhofer2021} and for the proposed Hybrid Model (HM). Where applicable, the notation remains consistent with that used in the body of the article.}
\begin{center}
    \footnotesize
    \begin{tabular}{l l  l l l}
    \toprule
    \textbf{Hyperparameter} &
    \textbf{Notation} &
    \textbf{LSTM} & \textbf{Transformer} & \textbf{Hybrid Model (HM)} \\
    \midrule    
    LSTM or Transformer Number of layers &  & 2 & 4 & 2 \\
    LSTM Hidden size & $h$ & 512 & 512 & 490 \\
    Initial embedding size & $z$ & - & 256 & 27 \\
    Reduced embedding size (after FFNN) & $z'$ & - & - & 6 \\
    Final MLP number of layers & & - & - & 2 \\
    FFNN inner hidden size & & - & 4096 & - \\
    Attention Heads & & - & 2 & - \\
    \midrule 
    Batch size &  & 128 & 128 & 128 \\
    Dropout probability & & 0.1 & 0.1 & 0.1 \\
    Embeddings dropout probability & & - & - & 0.4 \\
    Weight Decay &&  0.01 & 0.01 & 0.01 \\
    Learning rate & & 7e-5 & 7e-5 & 7e-5 \\
    Number of epochs & & 7 & 7 & 9 \\
    
    \bottomrule
    \end{tabular}
\end{center}
\end{table}

\section{Results on the test set}
\begin{table}[H]
\caption{Results on the test set for the LSTM baseline and the proposed model.}
\begin{center}
    \footnotesize
    \begin{tabular}{l l l l l}
    \toprule
    \textbf{Model} & \textbf{MAE} & \textbf{RMSE} & \textbf{F\textsubscript{1}} & \textbf{ROC-AUC}\\
    \midrule
    LSTM & 0.306 & 0.478 & 61.9 & 80.6 \\
    HM & 0.218  & 0.378 & 67.3 & 85.9 \\
    \bottomrule
    \end{tabular}
\end{center}
\end{table}

\section{5-Folds CV }
\label{app:tab:cv_results}
\begin{table}[H]
\caption{Extensive results of the 5-fold Cross-validation for the baseline LSTM and the proposed model.}
\begin{center}
    \footnotesize
    \begin{tabular}{l l l l | l l l}
    \toprule
    & \multicolumn{3}{c|}{\textbf{LSTM}} & \multicolumn{3}{c}{\textbf{HM}} \\
    \cmidrule {2-7}
    Fold & MAE & RMSE & F1 & MAE & RMSE & F1 \\
    \midrule
    1 & 0.347 & 0.553 & 58.34  & 0.244 & 0.433 & 60.22 \\
    2 & 0.365 & 0.570 & 42.79  & 0.302 & 0.519 & 59.67 \\
    3 & 0.272 & 0.444 & 66.22  & 0.254 & 0.404 & 75.22 \\
    4 & 0.332 & 0.548 & 44.82  & 0.266 & 0.433 & 59.84 \\
    5 & 0.310 & 0.504 & 63.88 & 0.299 & 0.502 & 71.06  \\
    \bottomrule
    \end{tabular}
\end{center}
\end{table}

\begin{table}[H]
\caption{Mean and standard deviation for each metric on the 5-fold Cross-validation results.}
\begin{center}
    \footnotesize
    \begin{tabular}{l l l l l}
    \toprule
    \textbf{Model} & \textbf{MAE} \scriptsize($\overline{x} \pm \sigma$) & \textbf{RMSE} \scriptsize($\overline{x} \pm \sigma$) & \textbf{F\textsubscript{1}} \scriptsize($\overline{x} \pm \sigma$) \\
    \midrule
    LSTM & 0.325 $\pm$ 0.036  & 0.524 $\pm$ 0.051 & 55.2 $\pm$ 0.108 \\
    HM & 0.273 $\pm$ 0.026 & 0.458 $\pm$ 0.050 & 65.2 $\pm$ 0.074 \\
    \bottomrule
    \end{tabular}
\end{center}
\end{table}

\section{Detail on the ablation study}
\label{tab:app:ablation_full}
\begin{table}[H]
\caption{Weekly results of the ablation study on the test set.}
\begin{center}
    \footnotesize
    \begin{tabular}{l l l l l l l l l l l l l}
    \toprule
    {} & \multicolumn{2}{l}{\textbf{Week 1}} & \multicolumn{2}{l}{\textbf{Week 2}} & \multicolumn{2}{l}{\textbf{Week 3}} & \multicolumn{2}{l}{\textbf{Week 4}} & \multicolumn{2}{l}{\textbf{Week 5}} & \multicolumn{2}{l}{\textbf{Week 6}} \\
    \cmidrule{2-13}
    Model & MAE & F\textsubscript{1} & MAE & F\textsubscript{1} & MAE & F\textsubscript{1} & MAE & F\textsubscript{1} & MAE & F\textsubscript{1}& MAE & F\textsubscript{1} \\
    \midrule
    HM & \textbf{0.126} & \textbf{82.2} & \textbf{0.169} & \textbf{74.7} & \textbf{0.209} & \textbf{68.6} & \textbf{0.244} & \textbf{64.0} & \textbf{0.269} & \textbf{58.6} &\textbf{0.294} & \textbf{51.0} \\
    TS+Att & 0.134 & 65.9 & 0.189 & 62.3 & 0.250 & 56.3 & 0.307 & 51.2 & 0.360 & 51.6 & 0.361 & 50.8 \\
    TS & 0.136 & 61.9 & 0.192 & 62.3 & 0.253 & 56.3 & 0.312 & 56.4 & 0.364 & 51.6 & 0.368 & 50.8 \\
    SF+TS & 0.144 & 73.9 & 0.203 & 62.3 & 0.262 & 56.3 & 0.320 & 51.2 & 0.374 & 49.3 & 0.375 & 50.8 \\
    SF & 0.779 & 20.4 & 0.746 & 22.7 & 0.752 & 25.3 & 0.713 & 18.9 & 0.754 & 19.5 & 0.787 & 17.0 \\
    \bottomrule
    \end{tabular}
\end{center}
\end{table}

\section{Location-agnosting vs location-specific training experiment}
\label{app:location_expe}

\begin{table}[H]
\caption{Weekly results for the HM on county vs national training data. The selected counties are Iowa (IA), Montana (MT) and Oklahoma (OK).}
\begin{center}
    \scriptsize
    \begin{tabular}{l l l l l l l l l l l l l l}
    \toprule
    {} & {} & \multicolumn{2}{l}{\textbf{Week 1}} & \multicolumn{2}{l}{\textbf{Week 2}} & \multicolumn{2}{l}{\textbf{Week 3}} & \multicolumn{2}{l}{\textbf{Week 4}} & \multicolumn{2}{l}{\textbf{Week 5}} & \multicolumn{2}{l}{\textbf{Week 6}} \\
    \cmidrule{3-14}
    Train & Eval. & MAE & F\textsubscript{1} & MAE & F\textsubscript{1} & MAE & F\textsubscript{1} & MAE & F\textsubscript{1} & MAE & F\textsubscript{1}& MAE & F\textsubscript{1} \\
    \midrule
    IA & IA & 0.101 & 86.7 & 0.179 & 67.7 & 0.214 & 69.5 & 0.287 & 63.5 & 0.298 & 60.9 & 0.272 & 59.4 \\
    MT & MT & 0.203 & 52.3 & 0.314 & 49.1 & 0.339 & 50.9 & 0.341 & 38.6 & 0.377 & 37.0 & 0.407 & 35.9 \\   
    OK & OK & 0.156 & 75.8 & 0.230 & 59.7 & 0.269 & 56.9 & 0.327 & 61.6 & 0.352 & 59.1 & 0.387 & 57.8 \\
    \midrule
      
    {} & IA & 0.086 & 89.3 & 0.122 & 79.2 & 0.151 & 78.4 & 0.189 & 71.9 & 0.214 & 73.9 & 0.235 & 66.4 \\
    all & MT & 0.144 & 59.5 & 0.168 & 52.9 & 0.178 & 50.0 & 0.209 & 45.5 & 0.237 & 44.4 & 0.265 & 38.6 \\   
    {} & OK & 0.096 & 83.1 & 0.160 & 75.8 & 0.196 & 77.9 & 0.209 & 77.5 & 0.260 & 73.5 & 0.298 & 66.3  \\
    \bottomrule
    \end{tabular}
\end{center}
\end{table}

\begin{table}[H]
\caption{Results on the test set using county vs national training data.}
\begin{center}
    \footnotesize
    \begin{tabular}{l l l l l}
    \toprule
    \textbf{Train} & \textbf{Eval.} & \textbf{MAE} & \textbf{RMSE} & \textbf{F\textsubscript{1}} \\
    \midrule
    IA & IA & 0.201 & 0.383 & 73.8 \\
    MT & MT & 0.301 & 0.354 & 46.7 \\  
    OK & OK & 0.278 & 0.402 & 63.1 \\
    \midrule
    {} & IA & 0.166 & 0.315 & 76.6 \\
    all & MT & 0.200 & 0.320 & 48.2 \\  
    {} & OK & 0.218 & 0.378 & 67.3 \\
    \bottomrule
    \end{tabular}
\end{center}
\end{table}

\end{document}